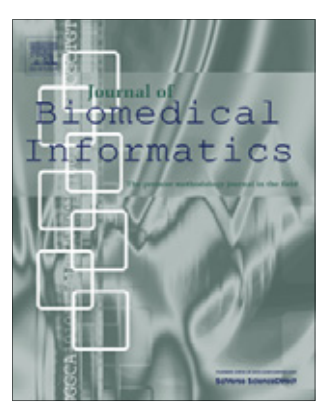

# Outlier detection for patient monitoring and alerting

Milos Hauskrecht [a,b,*], Iyad Batal [a], Michal Valko [a,c], Shyam Visweswaran [b,d], Gregory F. Cooper [b,d], Gilles Clermont [e]

[a] Department of Computer Science, University of Pittsburgh, Pittsburgh, PA, USA
[b] The Intelligent Systems Program, University of Pittsburgh, Pittsburgh, PA, USA
[c] Inria Lille-Nord Europe, equipe SequeL, Lille, France
[d] Department of Biomedical Informatics, University of Pittsburgh, Pittsburgh, PA, USA
[e] CRISMA Center, Department of Critical Care Medicine, University of Pittsburgh School of Medicine, Pittsburgh, PA, USA

ABSTRACT

We develop and evaluate a data-driven approach for detecting unusual (anomalous) patient-management decisions using past patient cases stored in electronic health records (EHRs). Our hypothesis is that a patient-management decision that is unusual with respect to past patient care may be due to an error and that it is worthwhile to generate an alert if such a decision is encountered. We evaluate this hypothesis using data obtained from EHRs of 4486 post-cardiac surgical patients and a subset of 222 alerts generated from the data. We base the evaluation on the opinions of a panel of experts. The results of the study support our hypothesis that the outlier-based alerting can lead to promising true alert rates. We observed true alert rates that ranged from 25% to 66% for a variety of patient-management actions, with 66% corresponding to the strongest outliers.

## 1. Introduction

Despite numerous improvements in health-care practice, the occurrence of medical errors remains a persistent and serious problem [1,2]. The well-known Institute of Medicine's report *To Err Is Human – Building a Safer Health System* estimated that between 44,000 and 98,000 Americans die each year as a result of medical errors [1]. The number of patients suffering morbidities due to such errors is estimated to be much higher. A 1999 study [3] estimates that the total national cost of injuries per year due to medical errors to be at least 17 billion dollars and that preventable injuries during hospital care affect 2% of hospital patients. More recent studies support that the actual rate of medical errors may be even higher than the above estimates [4]. A 2010 report [5] estimates that 13.5% of hospitalized Medicare beneficiaries experienced adverse events during their hospitalizations and that 44% of these events were preventable, yielding a medical error rate of about 6% ($0.135 \times 0.44$). A study of hospitals in North Carolina estimated that the medical error rate was between about 6% (external reviewers) and 16% (internal reviewers) [6].

The urgency and the scope of the medical errors problem have prompted the development of solutions to aid clinicians in eliminating such mistakes. Current computer tools for monitoring patients are primarily knowledge-based; the ability to monitor depends on the knowledge represented in the computer and extracted *a priori* from clinical experts. Unfortunately, these systems are time consuming to build and their clinical coverage is quite limited.

This paper presents a new data-driven monitoring and alerting framework that relies on stored clinical information of past patient cases and on statistical methods for the identification of clinical outliers (anomalies). It provides an expanded description of the methods and results originally reported in [7].

Our conjecture is that the detection of anomalies corresponding to unusual patient-management actions will help to identify medical errors. This *outlier-based (or anomaly-based) monitoring and alerting approach* can complement the use of knowledge-based alerting systems, thereby improving overall clinical coverage of alerting. In clinical sub-areas where knowledge-based alerting is not yet available, this new approach can serve as a standalone system.

Typical *outlier detection methods* identify unusual data instances that deviate from the majority of examples in the dataset [8,9]. In our approach the objective is different: we want to identify outliers in a given patient's care, where individual patient-management actions depend strongly on the condition of the patient. We build upon *conditional outlier detection methods* [7,10] to identify outliers in such settings. Our approach aims to identify patient-management actions for a given patient that are highly unusual with respect to past patients with condition(s) that are similar to the ones that the given patient suffers from. Once a deviation is detected, it may be used to generate a patient-specific alert for consideration by the clinician(s) who are caring for the patient.

* Corresponding author at: Department of Computer Science, 5329, Sennott Square, University of Pittsburgh, Pittsburgh, PA 15260, USA. Fax: +1 412 624 8854.
*E-mail address:* milos@pitt.edu (M. Hauskrecht).

Ideally we would like to have all statistical outliers correspond to medical errors. Hence their detection would lead to useful clinical alerts. However, in reality a patient-management action that is unusual from a statistical point of view does not always correspond to a helpful clinical alert. Nevertheless, our belief (and hypothesis) is that unusual patient-management actions that can be found by outlier detection methods will identify errors often enough to be worth alerting on. We report here our investigation of the relationship between conditional outliers and clinically useful alerts. We conducted experiments on data obtained from electronic health records (EHRs) of 4486 post-cardiac surgical patients and by having a subset of the 222 alerts generated by our approach evaluated by a panel of 15 experts in critical care medicine. Our results show that statistical outlier detection is positively correlated with clinically meaningful alerts, which provides support for our hypothesis that outlier-based alerting can be clinically useful.

## 2. Background

As a method of improving patient care, hospitals with EHRs often employ patient monitoring and alerting systems. These systems rapidly analyze patient data streams in order to identify events or conditions that may require the attention of clinical personnel. The notifications regarding the presence of these (typically adverse) events can come in the form of reminders or alerts. One important class of alerts aims to identify potential patient management errors. Examples include alerts that are designed to identify omission of an important medication, omission of an important laboratory test, or a prescription for a medication that is not consistent with a patient's condition and care.

Current computer systems that detect errors and alert on their occurrence are typically knowledge-based. Clinical knowledge acquired from domain experts is codified and represented (typically) as rules, which are then applied to patient data to detect adverse conditions or events [11,12]. Rule-based alerting systems have been developed for a range of medication decision support, such as drug-allergy checking, automated dosing guidelines, identifying drug-drug interactions, and detecting potential adverse drug reactions [13–16]. Such systems have also been developed for other clinical tasks, including monitoring of treatment protocols for infectious diseases and detection of deviations from such protocols [17], detection of growth disorders [18,19], and detection of clinically important events in the management of chronic conditions, such as diabetes [20] and congestive heart failure [21].

Rule-based systems have several advantages. They are based on clinical knowledge, and thus, are likely to be clinically useful. The rules are easy to automate and can be readily applied to patient data that are available in electronic form. However, rule-based systems suffer from several disadvantages. The creation of rules requires input from human experts, which can be tedious and time-consuming. Rules typically have limited coverage of the large space of potential adverse events, particularly more complex adverse events. Rule-based alerting systems are rigid and difficult to tune to achieve clinically acceptable performance in the environments in which they are deployed. It is not uncommon for alert rules to be retired (turned off) due to unacceptably high rates of false alerts [22]. Even when such rules remain active, the alerts generated by them may be ignored due to high false alert fatigue [23,24].

This paper describes the development and evaluation of a new type of alerting approach, one that is based on identifying clinical care outliers. This approach utilizes conditional outlier-based methods [7,10] to identify patient-management actions that are highly unusual with respect to past patients with same or similar conditions. Outliers thus represent actions that are unusual and *may* indicate patient-management errors. The advantages of the approach are that (1) it does not require expert input to develop a detection system, (2) clinically relevant outliers are derived empirically using a large set of prior patient cases and can be continually updated to reflect usual practice patterns, and (3) alert coverage can be broad and deep. Hence, this new approach has significant potential for wide, positive impact on clinical care.

## 3. Methods

Our outlier-based alerting approach consists of two stages: a *model-building stage* and a *model-application stage* (see Fig. 1). In the *model-building stage*, cases from the EHR repository are used to learn outlier models that summarize when (under what patient conditions) certain patient-management actions are typically made. This stage builds multiple outlier models to cover different patient-management actions, such as medication orders and laboratory test orders. For example, the heparin model captures patient subpopulations for which heparin is typically prescribed, subpopulations for which it is not, and subpopulations for which it is typically discontinued. In the *model-application stage* the outlier models are applied to new patient data to identify those actions that are unusual and deviate from the prevalent pattern of care, as represented in the outlier models. Details of the two stages are described in the following sections.

### 3.1. Conditional anomaly models

Anomaly detection is an active area of current machine learning and data mining research. An outlier (or a deviation or an anomaly) is an observation or a pattern in the data that appears to deviate significantly from other observations or patterns in the same data [8,9]. Anomaly detection methods have been applied to problems as diverse as monitoring of credit card transactions, detection of network intrusions, and detection of technical system failures.

Standard outlier detection methods try to identify unusual data instances [9]. In the clinical settings, these would correspond to unusual combination of symptoms, defining, for example, a rare disease. *Conditional anomaly detection (CAD)* [10] aims to detect unusual outcomes for a subset of (response) attributes given the values of the remaining (context) attributes. CAD is particularly suitable for detecting unusual outcomes, unusual behaviors, and unusual attribute pairings. In this work we use CAD to detect unusual patient-management actions in the context of an existing patient condition.

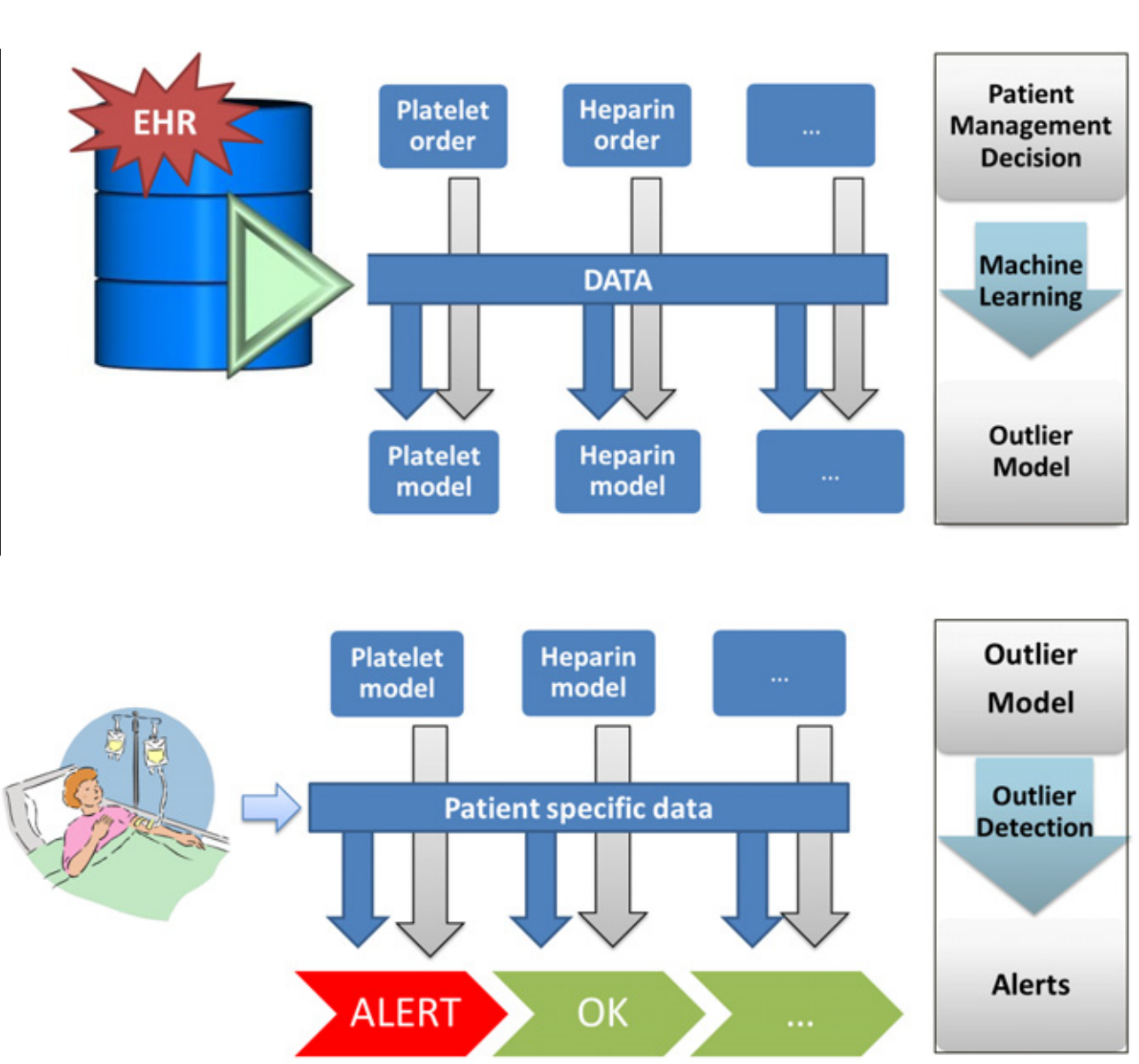

**Fig. 1.** Outlier-based alerting framework and its two stages. The model-building stage is shown on the top and the model-application stage is shown on the bottom.

*3.1.1. Measure of deviation*

In our approach, we quantify the deviation of a patient-management action using conditional probability measures. Let $y^*$ denote a patient-management action (such as a medication order) and let $x^*$ denote information about the current patient state. We say the action $y^*$ is *conditionally anomalous* given $x^*$, if the probability $P(y^*|x^*)$ is small, or equivalently if the probability $1 - P(y^*|x^*)$ is large. We represent the level of anomalousness of action $y^*$ for $x^*$ in terms of the following anomaly score:

$$\mathrm{Anom}(x^*, y^*) = 1 - P(y^*|x^*). \quad (1)$$

In practice the computation of this score consists of building a predictive model $P(y|x)$ for every patient-management action $y$ from past patient data, and its prospective application to observed $(x^*,y^*)$ pairs. Our approach for building a predictive model $P(y|x)$ consists of three steps: (1) segmentation and transformation of temporal data in the EHR, (2) representation of patient time-series data as fixed length vectors, and (3) learning of a probabilistic model $P(y|x)$. We describe each of these steps next.

*3.1.2. Segmentation of the EHR data*

The patient data in the EHR consist of complex multivariate time series combining results of laboratory tests, information on medication orders, procedures performed, diagnoses made, events encountered, and other information. In order to build an outlier detection model for patient-management actions, we first segment the data in every patient's EHR in time using discrete segmentation points. Time series data in the EHR observed up to each segmentation point time represent the patient state at that time. Each patient-state instance is then linked to a vector of patient-management actions that were made in between the current and the next segmentation point, which links the patient state with future actions following the state. Fig. 2 illustrates the process using 24-hour time segmentation, where the EHR for patient case A is segmented into four patient state instances that are linked to patient-management actions that follow the respective segmentation points.

*3.1.3. Representation of patient state and management actions*

Patient-state instances generated by the segmentation process cover different hospitalization periods and the amount of information in each of them may vary. To make the data and outlier models independent of these variations, we convert the time-series to a vector space representation of the patient state, where each state is defined using a fixed set of features and their values. We now briefly describe the features generated for laboratory tests, medication orders, and procedures.

- *Laboratory test features*: Laboratory tests with categorical values (e.g., with positive/negative values) are summarized using the following features: first value; second to last value; last value; time since last value; indicators of the test being performed and pending orders. Laboratory tests with continuous or ordinal values are summarized using 28 features that include the following: the difference between the last two measurements, the slope and the percentage drop in between them, features for nadir, apex, baseline values and their differences from last measured values. Nadir and apex values are the laboratory test values with the smallest and the greatest value recorded up to that point, respectively. Fig. 3 illustrates the definitions of some of these features for the platelet count laboratory test.
- *Medication order features*: Past and recent medications are summarized with four features per medication: (1) an indicator for whether the medication is active, (2) time since the first order of the medication, (3) time since the last order of that medication and (4) time since the last change in the medication.
- *Procedure features*: Procedure features capture information about procedures, such as heart-valve repair. We record three features per procedure: (1) an indicator of whether the procedure has ever been performed during the current hospitalization, (2) time since the procedure was first performed, and (3) time since the procedure was last performed.

In summary, the features in the vector-space representation abstract and cover multiple time-series for various clinical variables and events.

A predictive model $P(y|x)$ captures the relationship between the patient state $x$ and patient-management action $y$ that follows $x$. We consider two types of *patient-management actions*:

- Laboratory test orders with (true/false) values reflecting whether the laboratory test was ordered or not.
- Medication orders with (true/false) values reflecting if the medication was administered or not.

Hence all patient-management actions are represented using binary outcome variables linked to the corresponding patient-state vector.

*3.1.4. Learning predictive probabilistic models*

We construct the predictive model $P(y|x)$ for every action variable $y$ using the support vector machine (SVM) model [25]. More specifically, we use the libsvm library implementation of the SVM [26] and apply it using the linear kernel.

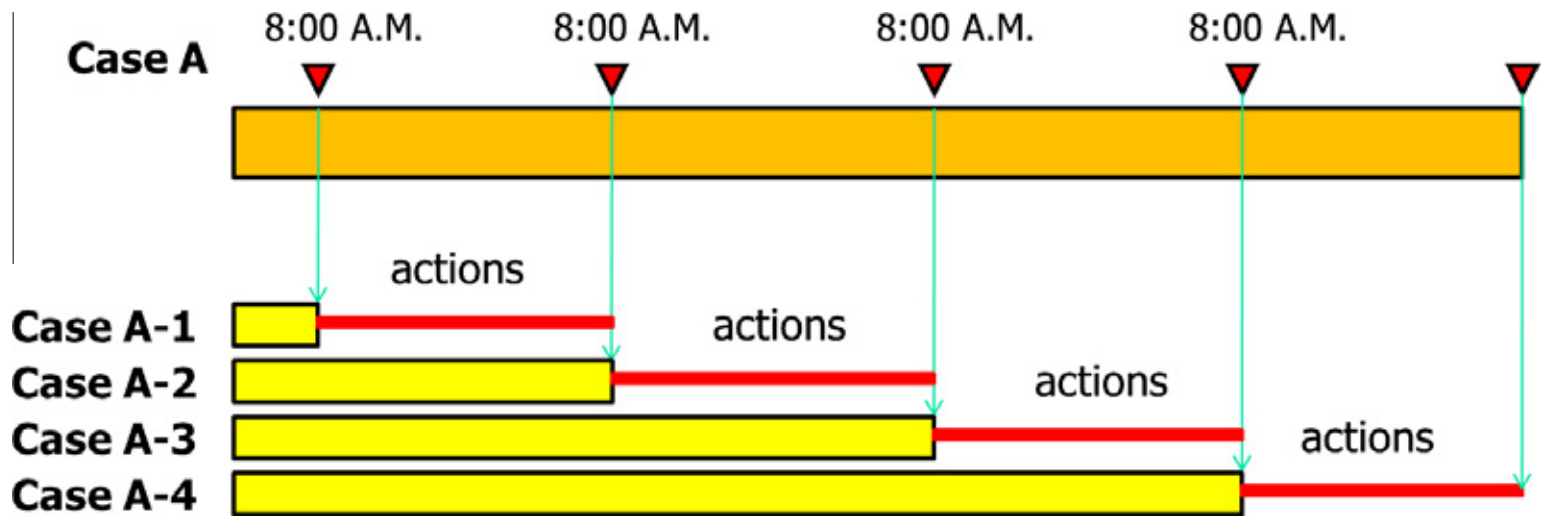

**Fig. 2.** The segmentation of a patient's EHR into four patient state – action instances.

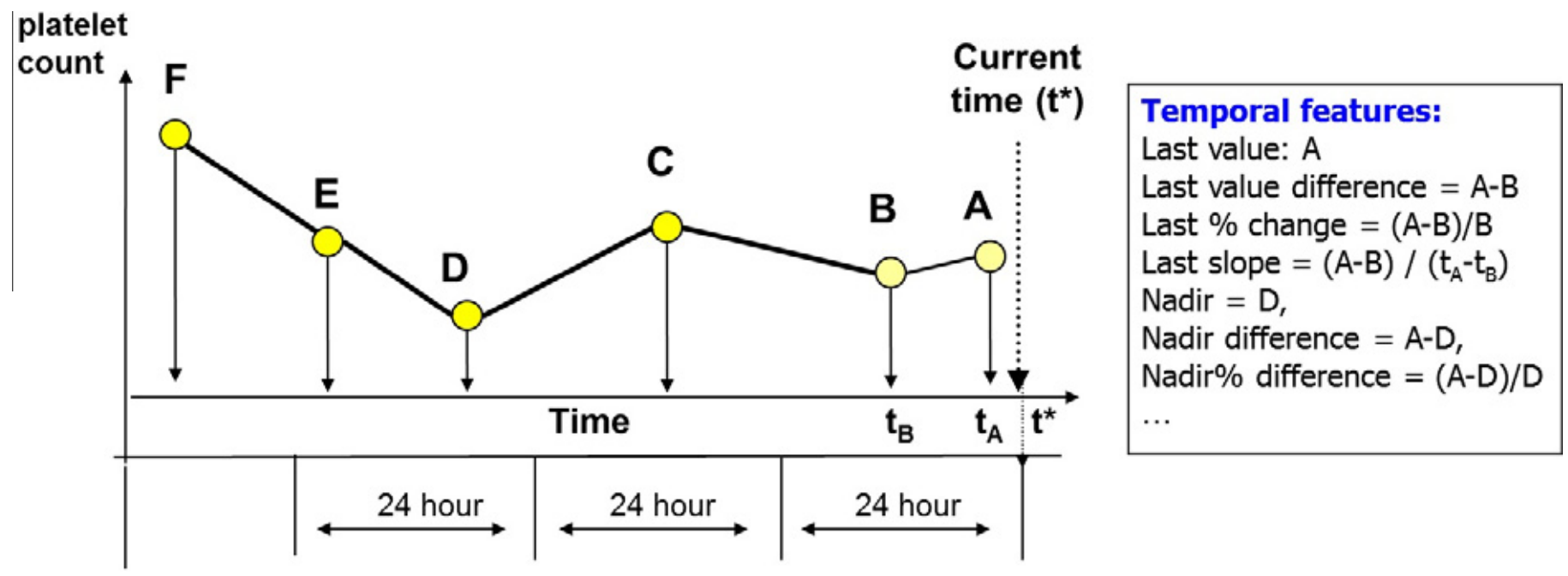

**Fig. 3.** Examples of temporal features for time-series of continuous laboratory test values.

The SVM is a discriminative learning model that is popular for learning high-quality discriminative patterns in high-dimensional datasets. It learns a discriminative projection $f(x)$ that aims to separate examples into two classes. Hence, it does not directly output a probabilistic model of $P(y|x)$. However, multiple probabilistic transformations of the SVM model are possible [27,28]. One such method was proposed by Platt [27]. The method works by refitting the discriminative projection induced by the SVM method using a logistic regression model. Unfortunately this approach is known to sometimes produce miscalibrated probabilistic estimates [29]. Hence, we developed a non-parametric approach to estimate the class posterior directly from data. Our method uses a discriminative projection $f(x)$ induced by the SVM. It uses $P(y = 1|f(x))$ to estimate the class posterior $P(y = 1|x)$ for any $x$, where $P(y = 1|f(x))$ is estimated non-parametrically from data and their projections.

The patient-state representation derived from the EHR includes thousands of features. It is not feasible to use all these features when learning predictive models. First, a particular patient-management action is likely to be influenced by only a relatively limited set of clinical variables (laboratory test values, medication orders) while other clinical information is often irrelevant for that action. Second, if all features are incorporated into the classification model via parameters, the high variance of these parameter estimates may negatively influence the quality of the model's predictions. Hence, a simpler model with fewer features and fewer parameters is often desirable. Our approach to addressing this problem is to build one model per action, and to select features for that model greedily and in groups. The feature groups are defined by temporal features characterizing the time-series of individual laboratory test values, medications, or procedures. For example, the platelet count feature group is formed by all temporal features representing the platelet count time-series. As another example, the heparin group consists of all features related to heparin orders. Briefly our algorithm works by first analyzing and assessing the predictive performance of individual feature groups. This is done by first learning a group-specific SVM classification model for each group and by assessing its performance on an internal training set in terms of the area under the Receiver Operating Characteristic curve (AUC) [30]. The groups are then combined into more complex predictive models by starting from the best group (the group with the highest AUC) and by considering the addition of the next best group (according to its initial AUC score) to the predictive model. A group is added, only if it improves the AUC performance of the model. The process stops when the top $k$ groups have been considered. In this study, we limit $k$ to 15. This value of $k$ was chosen for two reasons. First, it reflects our expectation the laboratory test and medication order patterns are in general sparse, that is, they are based on simple patterns that involve a small number of clinical variables and their features. Second, keeping $k$ small helps to speed up the model generation process.

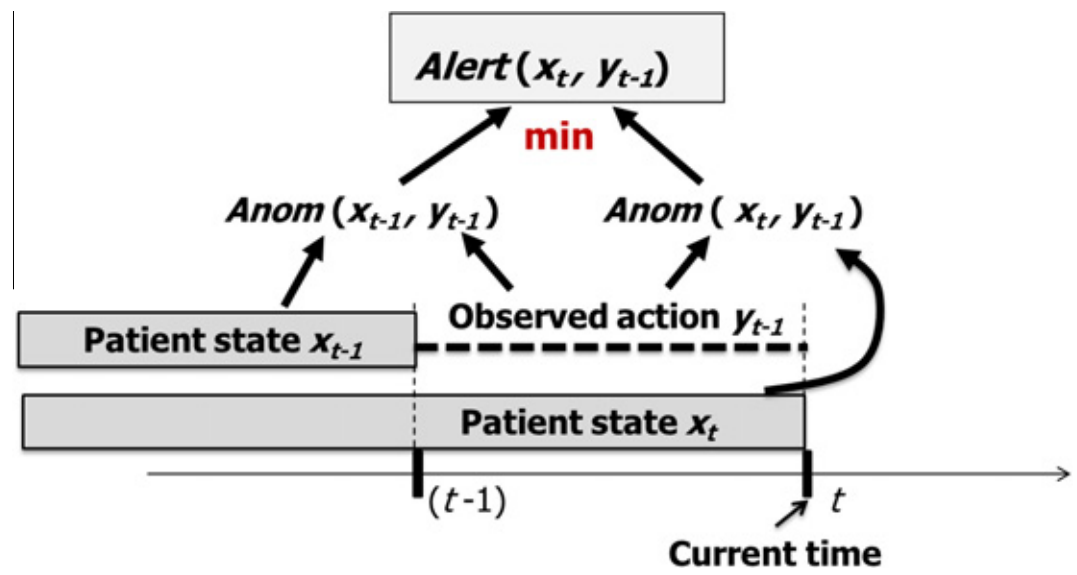

**Fig. 4.** Calculation of the alert score from the two anomaly scores.

### 3.2. The generation of alerts

A predictive model $P(y|x)$ described above captures the relation between the patient state $x$ and a patient-management action $y$. The model can be applied to a (new) patient instance $x^*$ and its associated action $y^*$ to calculate the anomaly score (as given by Eq. (1)) to measure how the actual patient-management action deviates from the predicted action. Our hypothesis is that these deviations are often clinically important and may correspond to patient-management errors; hence they are worthwhile to be alerted on. But how do we use anomaly scores to alert on data encountered prospectively?

Our alerting framework works by calculating an *alert score* that reflects the urgency (or severity) of the alert. The alert score is derived from two anomaly scores, as shown in Fig. 4. Let $x_t$ be the patient state instance at time $t$ (current time), $x_{t-1}$ be the patient instance in the previous time step $t-1$, and $y_{t-1}$ be a lab-order or a medication-order action executed in between times $t-1$ and $t$. We define the alert score for alerting on action $y_{t-1}$ given patient state $x_t$ at time $t$ as:

$$Alert(x_t, y_{t-1}) = min[Anom(x_{t-1}, y_{t-1}), Anom(x_t, y_{t-1})]. \quad (2)$$

Briefly, the first anomaly score reflects how surprising is the action taken over the most recent time window with respect to the previous patient state. The second score reflects how surprising that same action is in the next time step. The inclusion of the second anomaly score in the alert score simply assures that an unusual action taken recently remains unusual up to the current time. In other words the score reflects: (1) the degree of deviation of the action $y_{t-1}$ as defined by the anomaly score, and (2) the persistence of a high anomaly signal in two consecutive time steps: $t-1$ and $t$. If the *Alert* value (score) is above a threshold, an alert is generated at time $t$.

## 4. Evaluation of the outlier-based alerting methodology

### 4.1. Dataset

We evaluated our framework and its ability to generate clinically useful alerts using a dataset of 4486 post-surgical cardiac patients (PCP) [10,31] that was extracted from archived EHRs at a large teaching hospital in the Pittsburgh area. These EHRs were first divided into two groups: a training set that included 2878 cases seen in years 2002–2004, and a test set that included 1608 cases seen in years 2005–2006. Second, we used the time-stamped data in each EHR to segment the record at 8:00 A.M. every day to obtain multiple patient case instances, as illustrated in Fig. 2.

These patient instances were then converted into: (1) a vector-space representation of the patient state using the feature transformation described in the Methods section (Section 3.1.2), and (2) a vector representation of laboratory test-orders and medication actions with true/false values, reflecting whether the laboratory test or medication was ordered (and administered) within a 24-h period. The segmentation led to 51,492 patient-state instances, of which 30,828 were used for training and 20,664 were used for evaluation. The vector space representation of the patient state (at any point in time) included 9282 generated features for 335 laboratory tests, 407 medications, and 36 procedure categories. Rare laboratory tests, medications, and procedures (used in less than 20 patients) were excluded. Additional features we used were patient demographics (sex, age and race) and indicators of the presence of four heart-support devices. The patient-management action vectors linked with each patient-instance included 335 laboratory test orders and 407 medication orders.

### 4.2. Selection of models for evaluation

The training data were used to build three types of anomaly detection models: (1) models for detecting unexpected laboratory test-order omissions (lab-omissions), (2) models for detecting unexpected medication omissions (medication-omissions), and (3) models for detecting unexpected continuation of medications (medication-commissions). To build these models we used an SVM method with the greedy group-based feature-selection approach described in Section 3.1.4. To determine which models to include in the evaluation, a portion of the training data that was not used for model building and testing was used to derive the predictive performance of each model in terms of its AUC. Only predictive models with an AUC of 0.68 or higher were retained and applied on the test data; this selection of models was done to ensure that only models with moderate to strong predictive performance were used for anomaly detection. Table 1 shows the number of models with AUC of 0.68 or higher for each alert type.

These models were then applied to ⟨patient-state, action⟩ pairs in the test set to calculate the alert scores.

### 4.3. Selection of alerts for evaluation

The objective of our evaluation is to investigate whether outliers can lead to clinically useful alerts. In practice the alerting system would be implemented by setting a threshold on the anomaly (and/or alert score), that is, an alert would be raised if the alert score for the currently evaluated patient is greater than the threshold. To select ⟨patient-state, action⟩ alert candidates $\langle x_t, y_{t-1}\rangle$ for the evaluation study, we applied two action-specific thresholds. Briefly, for each action $a$, we define the minimum anomaly and the minimum alert score thresholds, denoted $anom(a)_{\min}$ and $alert(a)_{\min}$. These thresholds were used to filter ⟨patient-state, action⟩ alert candidates in the test set. The main motivation for applying these action-specific thresholds was to (1) reduce the size of the alert candidate pool for the study by focusing on more anomalous actions, and, at the same time, (2) assure good coverage of alerts on many different actions. More specifically, because of large observed differences in the anomaly and alert scores for different actions, applying global selection thresholds uniformly across all actions would have led to the selection of a relatively small subset of actions. Applying action-specific thresholds assured us of better action coverage in the alert candidate set.

The thresholds were set as follows. The threshold $anom(a)_{\min}$ for an action $a$, was obtained by (1) calculating anomaly scores $Anom(x_{t-1}, y_{t-1})$ for all observed ⟨patient-state, action⟩ pairs $\langle x_{t-1}, y_{t-1}\rangle$ in the test data for which $y_{t-1} = a$, and by (2) setting the threshold such that the top 125 observed $\langle x_{t-1}, y_{t-1}\rangle$ pairs with the highest $Anom(x_{t-1}, y_{t-1})$ score would pass it. Hence all alert candidates $\langle x_t, y_{t-1} = a\rangle$ for which either $Anom(x_{t-1}, y_{t-1})$ or $Anom(x_t, y_{t-1})$ score in Eq. (2) did not pass the $anom(a)_{\min}$ threshold were excluded. The alert score threshold $alert(a)_{\min}$, which was used to further filter the remaining $\langle x_t, y_{t-1} = a\rangle$ alert candidates, was set such that $alert(a)_{\min} \geqslant 0.15$; and at most the top 20 of these alerts, which are of the form $\langle x_t, y_{t-1} = a\rangle$, were used.

The application of the above threshold criteria on the test data resulted in 4870 alert candidates covering the most anomalous representatives of many different actions. From this set we selected 222 alerts on 100 patients. The selection of these 222 alerts from among the 4870 alert candidates was skewed towards alerts with higher alert scores to assure a good sample size and a more thorough analysis of stronger anomalies. Additional, alert selection was introduced by preferring alerts on patients with multiple alerts. This selection criterion was applied to make the patient case review process more efficient, because the reading of one patient case could support the evaluation of multiple alerts.

Fig. 5 shows the distribution of alert scores used for the evaluation (top) and the distribution of alert scores in the initial alert candidate set (bottom), by binning the alert scores in intervals of width of 0.2. Out of the total of 222 alerts, 101 alerts were lab-omission alerts, 55 were medication-omission alerts, and 66 were medication-commission alerts.

### 4.4. Review of alerts

The 222 alerts selected for evaluation were assessed by physicians with expertise in post-cardiac surgical care. There were 15 physician-reviewers of which 12 were fellows and 3 were faculty in the Departments of Critical Care Medicine or Surgery. The reviewers were given the patient cases and model-generated alerts for some of the patient-management actions, and were asked to assess the clinical usefulness of these alerts. The reviewers were divided randomly into five groups, with three reviewers per group, for a total of 15 reviewers. Overall, each clinician reviewed and assessed 44 or 45 alerts generated on 20 different patients. The reviews were conducted over the internet using secure web-based access to an interface developed by Post and Harrison [32].

*Assessment of alerts by reviewers*. For each alert, a reviewer was asked to:

**Table 1**
Models with AUC of 0.68 or higher.

| Alert type | Total number of models | Number of models with AUC $\geqslant$ 0.68 | Percentage of models retained |
|---|---|---|---|
| Lab omission | 335 | 197 | 58.80 |
| Medication omission | 407 | 278 | 68.30 |
| Medication commission | 407 | 231 | 56.75 |

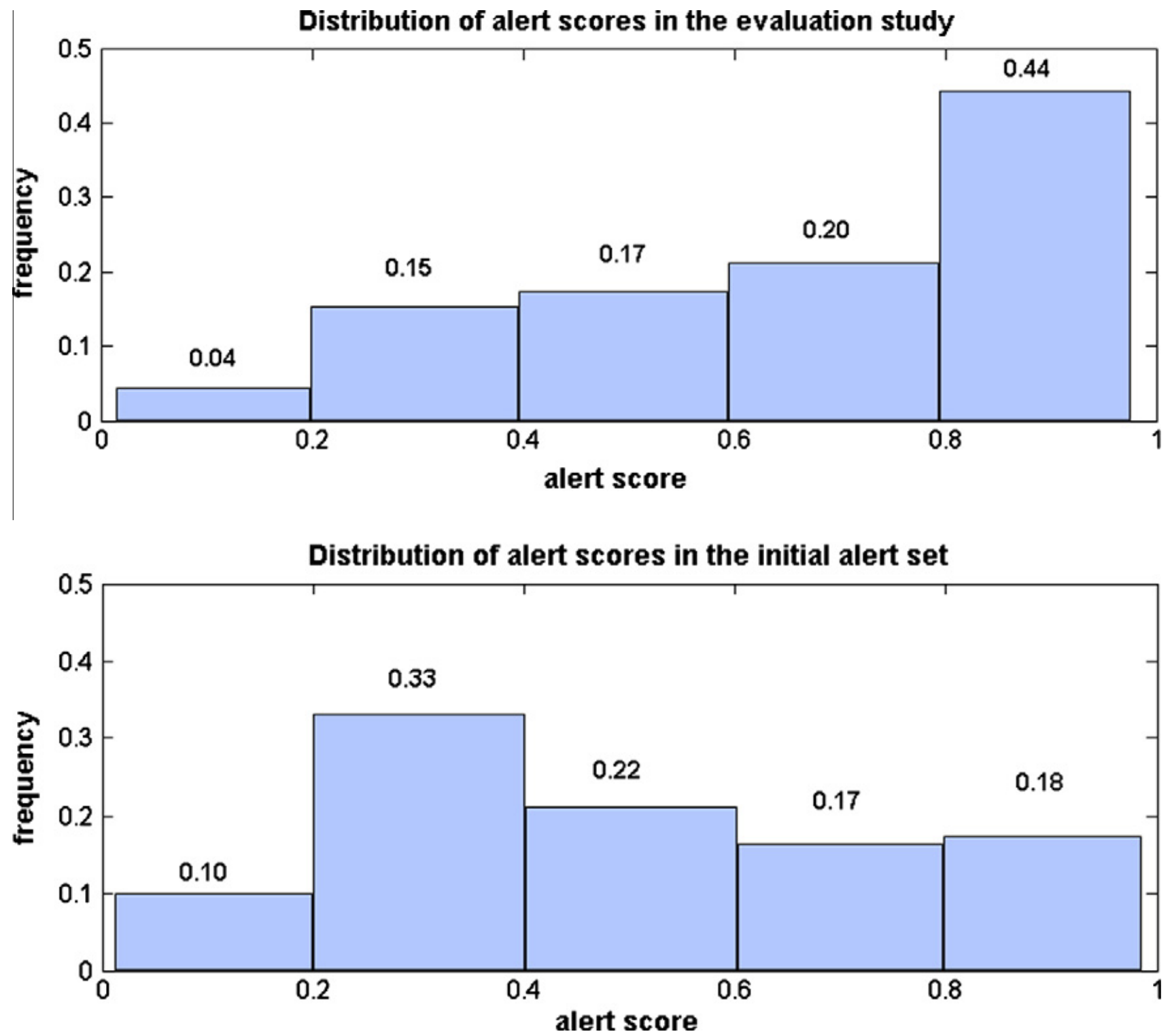

**Fig. 5.** Distributions of alert scores for 222 cases used in the evaluation (top panel), and alert scores for 4870 initially generated alert candidates (bottom panel).

*Item 1*. Assess the usefulness of the alert.
*Item 2*. Assess whether he or she would follow up the alert with a patient-management action.
*Item 3*. Provide free text comments and explanations justifying the decision for items 1 and 2.

To assess whether an individual alert was useful and whether it was likely to be followed up by an action, we used a simple majority rule. That is, an alert was considered to be useful if at least two out of three reviewers found it to be useful. Similarly, an alert was considered likely to be followed by an action if at least two out of three reviewers agreed that the alert should be followed up with a patient-management action.

## 5. Results

### 5.1. Reviewer agreement

Table 2 summarizes the pairwise reviewer agreements for the responses to the first two assessment items across groups in terms of the basic agreement (the fraction of cases in which two experts give the same answer) and Cohen's kappa statistics [33].

### 5.2. True alert rates

Table 3 summarizes the number of alerts, the number of correct alerts, and the true alert rates (together with 95% confidence intervals) based on responses to Items 1 and 2 for all alert cases used in the evaluation and for each alert category. Table 4 shows the same summary statistics for a subset of 'stronger' alerts that had alert scores in the [0.8, 1.0] range.

**Table 2**
Pairwise agreements among the reviewers in the five study groups. The minimum and maximum values of agreement and Cohen's kappa statistics between pairs of clinicians in each group are shown.

| Group | Item 1 | | Item 2 | |
|---|---|---|---|---|
| | Agreement min/max | Kappa min/max | Agreement min/max | Kappa min/max |
| 1 | 0.61/0.73 | 0.23/0.45 | 0.66/0.75 | 0.32/0.50 |
| 2 | 0.59/0.64 | 0.18/0.27 | 0.68/0.70 | 0.36/0.41 |
| 3 | 0.73/0.80 | 0.45/0.59 | 0.70/0.80 | 0.41/0.59 |
| 4 | 0.69/0.80 | 0.38/0.60 | 0.71/0.78 | 0.42/0.56 |
| 5 | 0.67/0.71 | 0.33/0.42 | 0.69/0.71 | 0.38/0.42 |

**Table 3**
True alert rates for Items 1 and 2 and their 0.95 confidence intervals.

| | Number of alerts | Item 1 | | Item 2 | |
|---|---|---|---|---|---|
| | | True alerts | True alert rate | True alerts | True alert rate |
| All | 222 | 121 | 0.55 [0.48; 0.62] | 114 | 0.51 [0.45; 0.58] |
| Lab omissions | 101 | 67 | 0.66 [0.56; 0.75] | 66 | 0.65 [0.55; 0.75] |
| Medication commissions | 66 | 32 | 0.49 [0.36; 0.61] | 30 | 0.46 [0.33; 0.58] |
| Medication omissions | 55 | 22 | 0.40 [0.27; 0.54] | 18 | 0.33 [0.21; 0.47] |

**Table 4**
True alert rates for Items 1 and 2 for stronger alerts with alert scores in the [0.8, 1.0] range and their 0.95 confidence intervals.

| | Number of alerts | Item 1 | | Item 2 | |
|---|---|---|---|---|---|
| | | True alerts | True alert rate | True alerts | True alert rate |
| All | 99 | 66 | 0.66 [0.56; 0.76] | 63 | 0.64 [0.53; 0.73] |
| Lab omissions | 70 | 51 | 0.73 [0.61; 0.83] | 50 | 0.71 [0.59; 0.82] |
| Medication commissions | 19 | 10 | 0.53 [0.29; 0.76] | 9 | 0.47 [0.24; 0.71] |
| Medication omissions | 10 | 5 | 0.50 [0.19; 0.81] | 4 | 0.40 [0.12; 0.74] |

### 5.3. Alert scores versus true alert rates

The alerts selected for the study belong to top action-specific anomalies and cover many patient management actions. However, since higher alert scores are associated with stronger statistical anomalies we expect them to also yield clinically useful alerts at a higher rate. This expectation is supported by entries in Tables 3 and 4 that list true alert rates for all alerts in the study, and a subset of alerts corresponding to stronger anomalies respectively.

Fig. 6 analyzes this relation (the relation of the alert score and the true alert rate) in more depth by binning the alert scores (in intervals of width of 0.2) and presenting the true alert rate per bin. The true alert rates for responses to Item 1 vary from 25% for low alert scores to 66% for high alert scores, indicating top action-specific alerts that come with higher alert scores are more likely associated with higher true alert rates. Similarly, alert rates for Item 2 range from 25% for the low alert scores to 64% for high alert scores. The increase in the true alert rate for higher alert score is also supported by the positive slopes of the lines in both figures that were obtained by fitting the results using linear regression.

In addition to the analysis in Fig. 6, we studied the relation of alert scores to true alert rates by analyzing how the alert scores induce true alert labeling in terms of the AUC score (the Wilcoxon–Mann–Whitney statistic). Briefly, all alerts reviewed in the study were ordered according to their alert scores. A true or false label was assigned to each alert based on whether the alert was found by the reviewers to be a true or a false alert. Note that this is different from the AUC of the predictive model (see Section 4.2.) that measures how well the model predicts a laboratory test or medication action observed in the health record. The AUC statistics for the alert score were 0.64 and 0.63 for Items 1 and 2, respectively. These are statistically significantly different from 0.5 (with the $p$-value 0.05), which is the value one expects to see for random or non-informative orderings. This again supports the finding that alerts with higher alert scores tend to induce better true alert rates.

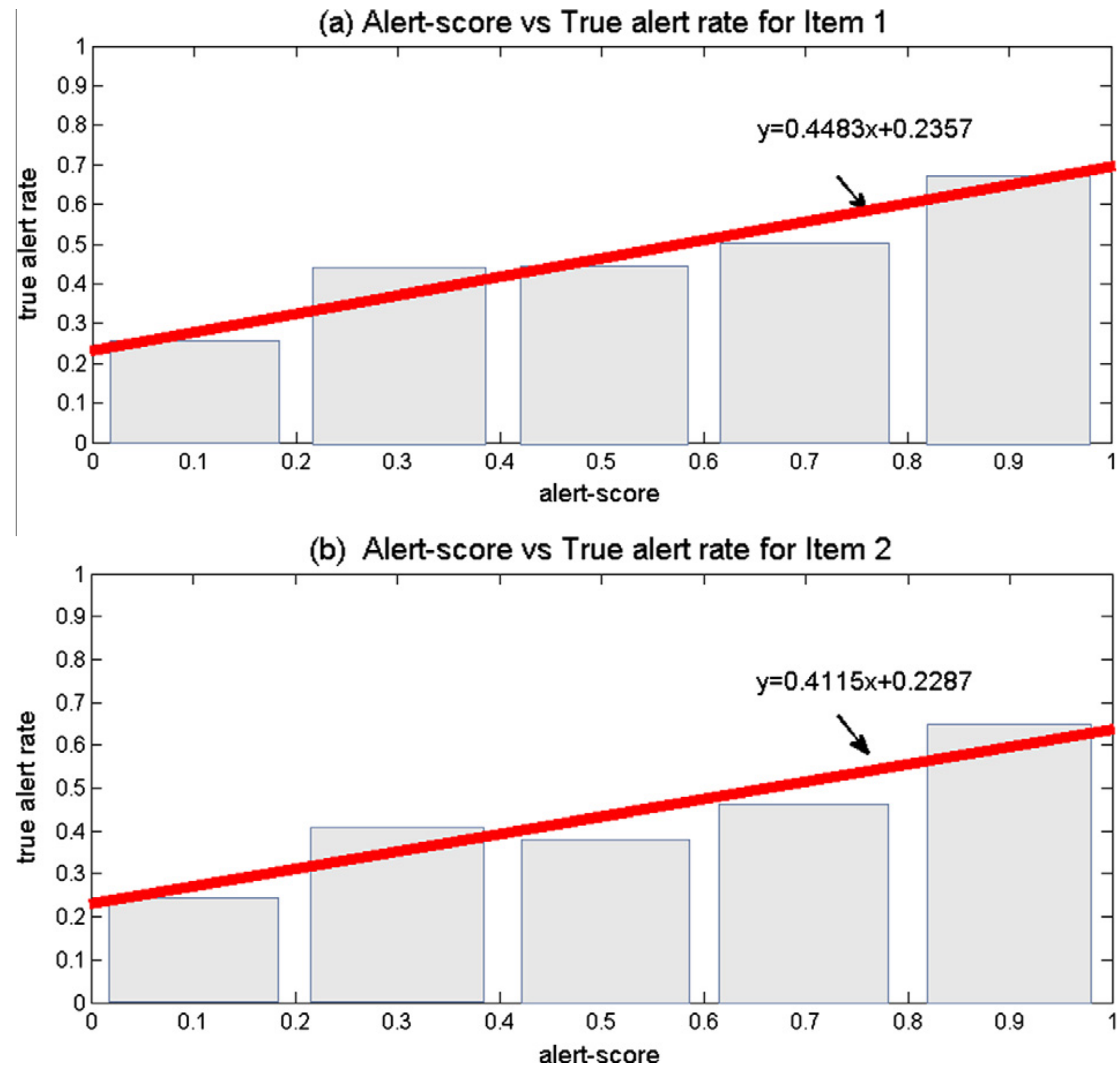

**Fig. 6.** The relation between the alert score and the true alert rate for responses to Item 1 (top) and Item 2 (bottom). The heights of the bins show true alert rates for alert-score intervals of width 0.2. The lines were fitted using linear regression.

### 5.4. Illustrative examples of alerts

To provide insight into how the above results were obtained, we present the following four illustrative examples of alerts. The first two examples were assessed as correct alerts, and the last two were assessed as incorrect alerts.

- *Alert 1. Order levothyroxine*: The patient was on levothyroxine prior to surgery. An order for one week of levothyroxine was sent to the pharmacy system. The patient eventually had to stay in the hospital longer but levothyroxine was not re-ordered. The system generated an alert and recommended re-ordering levothyroxine.
- *Alert 2. Order potassium*: The patient was in cardiogenic shock. The patient was on vasopressors and inotropes, as well as furosemide. The potassium levels were low. The system generated an alert and recommended supplementing potassium.
- *Alert 3. Order heparin*: The patient had undergone cardiac surgery 2 days ago and would, under normal circumstances, be given heparin after surgery. However, the patient was taken to surgery again for persistent postoperative bleeding at the time the alert was generated. This information was present only in the progress notes and was not available to the system; hence the system generated an alert and recommended continuing heparin.
- *Alert 4. Discontinue warfarin*: After heart valve replacement surgery, the patient was on heparin and was being transitioned to warfarin. The system generated an alert and recommended discontinuing warfarin. At the time of the alert, the INR (used to measure the intensity of anti-coagulation) was high, but not high enough for patients who have a mechanical valve.

## 6. Discussion

The experimental results reported above support that the proposed outlier-based methodology can generate clinically useful alerts with true alert rates ranging from 0.25 for weak alerts corresponding to weak outliers to 0.66 for stronger alerts.

Clinical alerting systems in the medical informatics literature are typically evaluated in terms of alert override rates that compare accepted and dismissed alerts [23,24,34,35]. The override rates may be influenced by multiple factors, such as the frequency (or the number) of the alerts and the quality of alerts [35–37]. In general, high frequency and low quality alerts can lead to alert fatigue and subsequently to high override rates [23,24,34–36]. In such a case, it is possible that overrides may include both unimportant and important (or useful) alerts. The override alert rates for a variety of drug safety systems reported in the literature were in the 0.49–0.96 range [23,24,34–36].

The analysis of override alert rates is typically associated with online deployed systems. Our evaluation study was conducted offline using retrospective data and hence it did not account for all aspects of the deployed alerting systems, such as alert fatigue, interruption of the workflow, and other factors. However, assuming that override rates of 0.49–0.96 in the drug safety literature closely approximate false alert rates, then the true alert rates reported in our study, which range from 0.25 (for weaker) to 0.66 (for stronger alerts), compare favorably to these numbers. Similarly, our results compare favorably to false alert rates of 0.86–0.99 reported for clinical monitoring systems in [38].

In addition, we showed that true alert rates are positively correlated with alert scores. This suggests the adjustment (control) of the alerting system toward desired true alert rates may be possible. Current approaches for controlling the false alert (or override rates) in the deployed systems include alert prioritization or alert tiering [36,39]. Other approaches include alert override prediction [40]. Our approach is a new direction for controlling alert system performance that uses empirically derived measures, rather than using annotated alert data.

In general, incorrect alerts identified during this study were due to information present only in text report data, interactions among medications, correlations among laboratory test orders that are part of the same panel, the use of less common alternatives to monitor or treat a patient, or the inability of the system to follow temporal patient management patterns (such as transition to warfarin from heparin in patients who have had valve replacement surgery). We believe it is likely that further improvements of the statistical models and the representation of additional information in the EHRs will lead to even higher true alert rates.

## 7. Conclusions

We proposed a new data-driven approach for identifying outliers in clinical care and for medical-error alerting. According to expert judgment, this approach is capable of identifying clinically valid outliers at reasonably high true positive alert rates across a broad range of clinical conditions and patient-management actions. The methodology has the advantage that it can be built directly from data in EHRs and does not depend on the construction of alerting models by experts. Hence it offers a complementary system to rule-based alerting systems for detecting potential medical errors.

A limitation of our evaluation study is that it was performed offline on retrospective patient cases outside of the regular clinical workflow and hence did not account for issues such as possible workflow interrupt and alert fatigue. Another, limiting factor of the study is the number of patient cases and alerts the expert reviewers could assess, which prevented us from studying many different aspects of the outlier-based alerting system. First, in order to study alerts across a broad range of actions, we limited the alert instances for each action to be among the top 125 anomalies and the top 20 alerts for that action. Studying a broader range of anomalies and alerts for each individual action could provide additional insights on how to better control action-specific alerts and their alert rates. Second, the skewed selection of alerts towards higher alert scores and towards patients with a larger number alerts (to make the review process more efficient) could also influence the obtained results. For example, patients with multiple alerts may be more likely to be hospitalized for longer periods of time and consequently may represent more complicated cases. We plan to address these limitations in future studies of the approach.

Finally, we believe many improvements of this new approach are possible. First, our current outlier models are built for each patient management action individually and do not consider multivariate relations among them; hence they may fail to identify multivariate outliers. For example, an outlier identified in the current medication order may ignore other alternative medications that could be used to substitute for it, and hence the outlier model may incorrectly generate an omission alert if the medication was substituted by an alternative. Second, our current method relies on the time segmentation of 24 h that was used both for segmenting the patient record into patient instances and for predicting actions. This segmentation resolution may lead to reduced ability to reliably distinguish and predict orders of medications with shorter time-horizon effects, such as epinephrine that is given in acute care life-support settings. Third, while our current predictive features were very good for many actions, they still failed to yield good predictive models for others. The investigation of new, more suitable feature sets that characterize complex time-series data may lead to further improvements and better coverage of patient management actions with such models.

## Acknowledgments

This research work was supported by Grants R21-LM009102, R01-LM010019, and R01-GM088224 from the NIH. The work was done entirely at the University of Pittsburgh. Its content is solely the responsibility of the authors and does not necessarily represent the official views of the NIH.